\PassOptionsToPackage{numbers,sort&compress}{natbib}
\documentclass[10pt,a4paper,logo]{paper}
\usepackage{natbib}
\ifdefined\pdfsuppressptexinfo\pdfsuppressptexinfo=-1\relax\fi
\usepackage{algorithm,algpseudocode,multirow,subcaption}
\definecolor{title_blue}{HTML}{204899}
\definecolor{cite_blue}{HTML}{044DC1}
\definecolor{cite_purple}{HTML}{7406A7}
\newcommand{\ours}{\textup{\textsc{RePace}}}
\hypersetup{citecolor=cite_blue,linkcolor=cite_purple,urlcolor=cite_purple,
  pdftitle={RePace: Reasoning about Demonstration Speedup for Faster Vision-Language-Action Policies},
  pdfauthor={Insup Lee, Hojin Jeon, Sunguk Shin, Seoungbin Bae, Taekwan Kim, Jinwoo Shin, Dongsu Han}}
\title{\ours{}: Reasoning about Demonstration Speedup for Faster Vision-Language-Action Policies}

\author[1,*]{Insup Lee}
\author[1,*]{Hojin Jeon}
\author[2]{Sunguk Shin}
\author[1]{Seoungbin Bae}
\author[2]{Taekwan Kim}
\author[1,3]{Jinwoo Shin}
\author[1]{Dongsu Han}
\affil[1]{KAIST}
\affil[2]{Korea University}
\affil[3]{RLWRLD}
\renewcommand{\RePaceAuthorLegend}{{\footnotesize\textsuperscript{*}Equal contribution.\par}}

\begin{abstract}

Since vision-language-action (VLA) policies can inherit the slow pace of human demonstrations, accelerating these demonstrations before fine-tuning can lead to faster VLA policies without an additional speed-selection module at deployment.
Several offline speedup methods have focused on identifying segments that require less precision and assigning predefined speedup factors to the segments.
However, even within a single segment, a speedup factor suitable for steady motion may be too high when precise alignment is required.
To this end, this paper presents \ours{}, an offline reasoning framework that determines how much to accelerate each part of a demonstration.
For the same recorded movement, \ours{} measures how much each candidate speedup factor shortens the sequence and compares the resulting observation-action pairs used for training.
Given the task instruction, a pretrained language model uses the \textit{visual context} and \textit{original motion} of successive windows to interpret these \textit{candidate retiming effects} and select a speedup profile for the complete episode.
Experimental results demonstrate that \ours{} outperforms the baselines in overall task execution efficiency on both LIBERO and RoboCasa-Kitchen.
On all three tasks with a gripper-equipped Franka Research 3 arm, \ours{} matches or exceeds the acceleration baselines in success rate and completes successful trials faster.

\end{abstract}

\begin{document}
\maketitle
\section{Introduction}
\label{sec:intro}

Vision-language-action (VLA) models extend pretrained vision-language representations to robot action prediction~\cite{brohan2023rt}, and researchers have evaluated these models on various real-world manipulation tasks~\cite{ma2026survey}.
A pretrained VLA policy can be fine-tuned on demonstrations of a new task to predict robot actions from observations and a language instruction~\cite{kim2024openvla}.
Practical deployment of these policies requires both successful manipulation and timely task completion.
However, since even experienced demonstrators may move the robot cautiously to avoid mistakes during teleoperation~\cite{zhong2025survey}, a policy trained on these demonstrations can inherit their slow pace, even in open space where the robot could move faster.

\begin{figure}[H]
  \centering
  \includegraphics[width=0.70\linewidth]{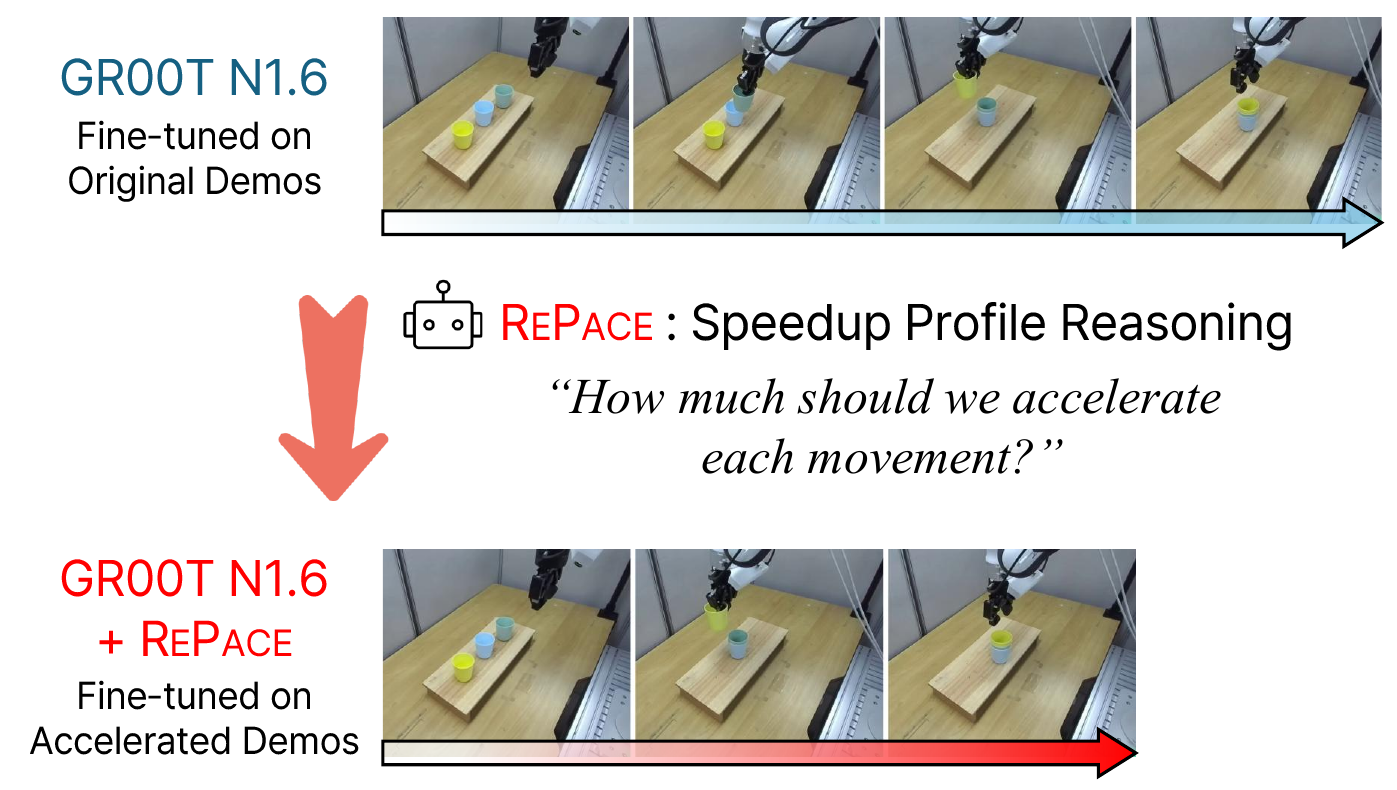}
  \caption{Offline demonstration acceleration with \ours{}.
  \ours{} compares how candidate speedup factors change the observations and action targets used for VLA fine-tuning.
  It uses the recorded scene and motion to interpret these changes and select a speedup profile.
  }
  \label{fig:teaser}
\end{figure}

For faster VLA policy, there have been two primary approaches to accelerate policies at different stages: (i) test-time policy acceleration and (ii) offline demonstration acceleration.
Although several studies have focused on acceleration during execution (i.e., test-time policy acceleration)~\cite{arachchige2025sail,kim2026time,jeon2026pacevla,hu2026freespeed,shirasaka2026robopace,wu2026speedup}, the task success rate can decline when predicted action chunks are directly downsampled at higher speedup factors~\cite{guo2025demospeedup}.
Meanwhile, other studies have addressed the acceleration of demonstrations before policy training (i.e., offline demonstration acceleration)~\cite{guo2025demospeedup,kim2026espada,sanchez2026volt,cao2026adatempo,oh2026egospeedup,moon2026retiming}.
Although the overhead of VLA fine-tuning makes it difficult to use downstream policy performance to guide demonstration acceleration, we focus on the latter approach, \textit{offline demonstration acceleration}, because it enables faster VLA policies without a speed-selection module at deployment.
Several studies have addressed the question of where to accelerate demonstrations offline.
DemoSpeedup~\cite{guo2025demospeedup} uses action entropy to identify segments with different precision requirements, while VOLT~\cite{sanchez2026volt} uses a vision-language model (VLM) to interpret visual context for segment selection and assigns a human-defined speedup rate.

Even within a segment selected for acceleration, we still need to decide \textit{how much to accelerate each movement}.
For example, suppose that we treat the entire approach to a cup before grasping as a single segment.
The gripper may move straight toward the cup and then shift slightly to one side to position its fingers for the grasp.
A lower speedup factor may preserve the observation at this change in direction, while a higher factor may combine successive commands and omit that observation.
Our key insight is that, before policy fine-tuning, we can compare candidate factors directly by examining how many steps each saves and how each changes the observations and commands used for learning.
In consideration of these changes, \textit{how should the recorded scene and motion guide the choice of speedup factor throughout a demonstration?}
In addressing this question, we have derived the following three challenges in offline speed selection.

\textbf{Challenge 1) How can we compare the effects of candidate speedup factors before policy fine-tuning?}
A shorter demonstration may omit details of the movement that the policy needs to learn.
The choice between two speedup factors therefore depends on how many additional steps the higher factor saves and which details the lower factor preserves.

\textbf{Challenge 2) How can we determine which data changes matter for the task?}
The significance of a change in the training data depends on the scene and the recorded motion.
For example, a few observations may capture a straight path through open space, while an intermediate observation can show how the gripper adjusts an object to fit through a narrow opening.

\textbf{Challenge 3) How should the surrounding movements guide the choice of speedup for each window?}
A short window of consecutive demonstration steps may show how an object moves without revealing the purpose of that movement.
For example, seeing the object placed in a holder in the next window can help explain why a small rotation just before placement may warrant a lower speedup factor.

To this end, we propose \ours{}, an offline framework that selects a speedup profile for the complete episode, as shown in Fig.~\ref{fig:teaser}.
We measure the effects of each candidate factor under the same retiming rules used to construct the accelerated demonstrations.
A pretrained language model~\cite{qwen3.8} interprets these measurements considering the task instruction, scene descriptions, and recorded motion from successive windows to select the profile without requiring window-level speed labels.
Our main contributions are summarized as follows.

\begin{itemize}
  \item We propose \ours{}, an offline demonstration acceleration framework that uses large language model (LLM)-based reasoning to select speedup factors and construct accelerated demonstrations.

  \item We define three inputs for effective speedup reasoning: \textit{visual context} and \textit{original motion} describe the recorded demonstration, while \textit{candidate retiming effects} quantify the changes produced by each candidate speedup factor.

  \item We evaluate \ours{} using both simulation benchmarks and a real-world Franka Research 3 robot, showing that demonstrations accelerated via \ours{} contribute to faster VLA policies while mitigating decreases in task success rates.
\end{itemize}

\section{Related Work}
\label{sec:related}

This section briefly reviews the previous work on vision-language-action models and the acceleration methods for faster robot policies.

\subsection{Vision-Language-Action Models}

Given visual observations and a task instruction, VLA models can predict discrete action tokens~\cite{brohan2023rt} or generate continuous action sequences through flow matching~\cite{gr00tn1_2025,black2024pi_0,intelligence2025pi_}.
Robot policies such as Diffusion Policy (DP)~\cite{chi2025diffusion} and Action Chunking with Transformers (ACT)~\cite{zhao2023learning} predict several actions at a time as an action chunk.
VLA models can encode such chunks as compact token sequences~\cite{pertsch2025fast} or decode multiple actions in parallel~\cite{kim2025fine}.
To determine which actions a task requires, some methods translate instructions into control programs~\cite{liang2023code} or construct spatial maps to guide robot motion~\cite{huang2023voxposer}.
To guide action prediction, some VLA policies describe the steps needed to follow a task instruction or predict an image of the scene they aim to reach~\cite{zawalski2024robotic,zhao2025cot,li2026trm}.
For example, a predicted image may show where to place an object without specifying how quickly the gripper should move it there.

\subsection{Robot Policy Acceleration}
Several studies have addressed test-time policy acceleration by controlling how quickly a robot executes its movements~\cite{arachchige2025sail,kim2026time,jeon2026pacevla,hu2026freespeed,shirasaka2026robopace,wu2026speedup}.
For example, SAIL~\cite{arachchige2025sail} adjusts the interval between actions according to motion complexity or gripper transitions.
Related work on the temporal structure of robot actions has explored variable sampling frequencies and waypoint-based representations~\cite{masuya2025variable,shi2023waypoint,belkhale2023hydra}.
Speed selection can also enter policy training. TempoVLA~\cite{jing2026tempovla} trains a policy whose speed can be specified at execution time, while AutoSpeed~\cite{hu2026autospeed} selects retimed targets during training based on prediction error and horizon.

Offline methods resample recorded demonstrations before policy training~\cite{guo2025demospeedup,kim2026espada,sanchez2026volt,yang2026improving}.
For example, DemoSpeedup~\cite{guo2025demospeedup} and VOLT~\cite{sanchez2026volt} identify segments for acceleration using action entropy and visual task context, respectively.
Other methods retime demonstrations by comparing the timing of similar movements in robot trials~\cite{cao2026adatempo}, transferring the timing of human movements to robot demonstrations~\cite{oh2026egospeedup}, or predicting speed along a recorded path~\cite{moon2026retiming}.
However, the cues used in these offline demonstration acceleration studies~\cite{guo2025demospeedup,kim2026espada,sanchez2026volt,cao2026adatempo,oh2026egospeedup,moon2026retiming} do not by themselves show how different speedup factors would change the observations and action targets for the same movement.
This motivates the design of \ours{}, which compares these changes in the context of the recorded scene and motion to select a speedup profile (Table~\ref{tab:prior_speed_selection}).

\begin{table}[t]
  \centering
  \caption{Comparison of robot policy acceleration methods.}
  \label{tab:prior_speed_selection}
  \fontsize{8.5}{10}\selectfont
  \renewcommand{\arraystretch}{0.8}
  \setlength{\tabcolsep}{2pt}
  \renewcommand{\arraystretch}{1.08}
  \begin{tabular}{@{}p{0.24\columnwidth}p{0.20\columnwidth}p{0.525\columnwidth}@{}}
    \toprule
    \raggedright Method & \raggedright Stage & \raggedright Basis for speed assignment \tabularnewline
    \midrule
    \raggedright SAIL~\cite{arachchige2025sail} & \raggedright Test-time &
    \raggedright Motion complexity or gripper transitions \tabularnewline
    \raggedright TempoVLA~\cite{jing2026tempovla} & \raggedright Training, Test-time &
    \raggedright Speed sampling in training, user or VLM control at execution \tabularnewline
    \raggedright AutoSpeed~\cite{hu2026autospeed} & \raggedright Training &
    \raggedright Prediction error and horizon of retimed targets \tabularnewline
    \raggedright DemoSpeedup~\cite{guo2025demospeedup} & \raggedright Offline &
    \raggedright Predefined rates for segments identified by action entropy \tabularnewline
    \raggedright VOLT~\cite{sanchez2026volt} & \raggedright Offline &
    \raggedright User-set factor for segments labeled by a VLM \tabularnewline
    \raggedright \ours{} (Ours) & \raggedright Offline &
    \raggedright Candidate retiming effects in visual and motion context \tabularnewline
    \bottomrule
  \end{tabular}
\end{table}

\section{Method}
\label{sec:method}
This section describes the problem setup, the design principles of \ours{}, and its architecture.

\subsection{Problem Setup}
\label{sec:problem}
We consider the offline acceleration of demonstrations for fine-tuning a pretrained VLA policy.
A demonstration $\tau$ associated with a task instruction $\ell$ is defined as follows:
\begin{equation}
    \tau=\{(o_t,q_t,a_t)\}_{t=1}^{T},
  \label{eq:research_demonstration}
\end{equation}
where $o_t$, $q_t$, and $a_t$ denote the observation, recorded robot state, and action at step $t$, respectively.
We divide each demonstration into ordered, non-overlapping windows $\mathcal{W}_1,\ldots,\mathcal{W}_W$ and assign each window a speedup factor $r_w$ from a finite candidate set $\mathcal{R}$.
Fixed boundaries let us compare candidates over the same interval without phase annotations or a segmentation model.
When $1\in\mathcal R$, the speedup profile also determines which windows keep their original timing.
During retiming, we resample observation-action pairs and transform commands according to the action representation.
For a speedup profile $\mathbf{r}=(r_1,\ldots,r_W)$, the accelerated demonstration $\widetilde{\tau}$ is defined as follows:
\begin{equation}
  \widetilde{\tau}=\Phi(\tau,\mathbf{r}),
  \label{eq:retiming}
\end{equation}
where $\Phi$ denotes the retiming operator that groups consecutive steps within each window according to the selected factor.
We fix the action representation, sampling rules, command transformation, and operator settings before comparing factors.

We construct the accelerated dataset $\widetilde{\mathcal{D}}$ by applying Eq.~\ref{eq:retiming} to each demonstration in the source dataset $\mathcal{D}$ with its selected speedup profile.
We fine-tune a VLA policy on $\widetilde{\mathcal{D}}$ and use a policy fine-tuned on the original demonstrations as the baseline.
Our goal is to shorten task execution while mitigating the decline in task success rate that can accompany acceleration.
To select a profile before fine-tuning, \ours{} compares how candidate speedup factors change the training data for the same recorded movement.

\subsection{\ours{} Overview}
\label{sec:overview}
\ours{} follows three design principles when comparing candidates.
(i) We measure actual step reduction and changes to observation-action pairs under the same retiming rules used for data construction.
(ii) The selector interprets these changes alongside recorded visual context and motion.
(iii) The selector considers the complete episode when assigning factors to individual windows.
To interpret the numerical measurements in the context of the task instruction and scene descriptions, we use a pretrained language model without further training or window-level speed labels.

\begin{figure}[t]
  \centering
  \includegraphics[width=0.95\textwidth]{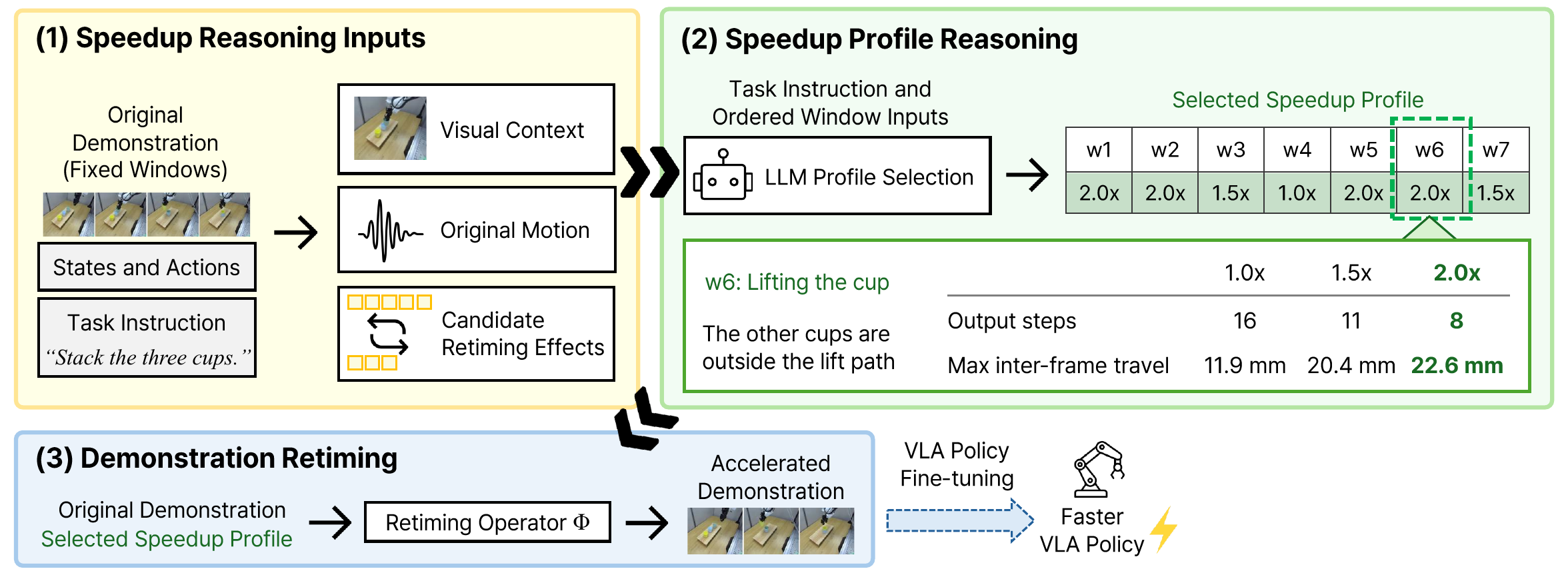}
  \caption{Architecture of \ours{}: (i) \textit{Speedup reasoning inputs} describe visual context, original motion, and candidate retiming effects for each window; (ii) \textit{Speedup profile reasoning} selects a profile for the complete episode from the task instruction and ordered window inputs; (iii) \textit{Demonstration retiming} applies the selected profile with the same operator and settings used to compute candidate effects.}
  \label{fig:method_overview}
\end{figure}

As illustrated in Fig.~\ref{fig:method_overview}, \ours{} processes each demonstration through three components.
\textbf{Speedup reasoning inputs} distinguish \textit{visual context} and \textit{original motion} from \textit{candidate retiming effects}, which describe how each factor changes the training data.
In \textbf{speedup profile reasoning}, the model selects a full-episode profile from the instruction and ordered window inputs and identifies the evidence for its choices.
\textbf{Demonstration retiming} applies the selected profile under the rules used to compute candidate effects, producing observation-action pairs for policy fine-tuning.
Sections~\ref{sec:inputs}, \ref{sec:reasoning}, and \ref{sec:retiming} describe these components, respectively.

\subsection{Speedup Reasoning Inputs}
\label{sec:inputs}
For each source window $w$, the three reasoning inputs are defined as follows:
\begin{equation}
  x_w=\big(v_w,m_w,\{e_w(r)\}_{r\in\mathcal{R}}\big),
  \label{eq:inputs}
\end{equation}
where the visual context $v_w$ describes the spatial relations relevant to the task and the original motion $m_w$ summarizes how the gripper moves in that scene.
For the same recorded movement, the candidate retiming effects $e_w(r)$ quantify how each speedup factor $r$ changes the observation-action pairs under $\Phi$.
Appendix~\ref{app:cue_definitions} describes how we extract the three inputs, and Table~\ref{tab:numerical_input_mappings} in Appendix~\ref{app:platform_rules} lists the action representations and unit conversions for each platform.

\Needspace{3\baselineskip}
\noindent\textbf{Visual context.}
A vision-language model (VLM) examines the recorded frames to describe the arrangement of objects identified from the task instruction, obstacles along the path, and the space available for movement.
The description includes frame references for visible relations and marks unobservable relations as unknown.

\smallskip
\Needspace{3\baselineskip}
\noindent\textbf{Original motion.}
In similar visual settings, motion measurements help distinguish steady transport from small adjustments in position and orientation.
From the recorded robot states, we measure end-effector displacement, path length, and speed, as well as angular path length and maximum angular speed.
Changes in finger aperture describe the measured finger motion, while gripper-command transitions indicate when the robot requests opening or closing.
For each window, we record the transitions within it and the nearest transitions before and after it.

\smallskip
\Needspace{3\baselineskip}
\noindent\textbf{Candidate retiming effects.}
For each candidate, we count the output steps under the sampling and gripper rules.
When comparing adjacent factors, we also count how many additional steps the higher factor saves.
We measure peak command norms after retiming because merging successive commands can increase the displacement requested in a single control step.
The maximum translational and angular path lengths between successive selected observations capture intervening movement that endpoint displacement can miss.
For each source gripper-command transition, we identify the corresponding output step.
Appendix~\ref{app:candidate_effects} defines these measurements and the criteria for output equivalence.

\Needspace{7\baselineskip}
\subsection{Speedup Profile Reasoning}
\label{sec:reasoning}
A pretrained language model selects a speedup profile for the complete episode from the task instruction and the ordered window inputs defined in Eq.~\ref{eq:inputs}.
We express this selection as
\begin{equation}
  (\mathbf{r},E)=f_{\mathrm{LLM}}(\ell,x_1,\ldots,x_W;\mathcal{I}),
  \label{eq:selector}
\end{equation}
where $\mathcal I$ specifies how to compare candidate speedup factors and $E$ contains brief explanations with references to the inputs.
The retiming operator $\Phi$ computes the changes to the training pairs for each candidate factor, and the model interprets these changes in the context of the recorded scene and motion.

\smallskip
\Needspace{3\baselineskip}
\noindent\textbf{Joint cue interpretation.}
In the cup example from Section~\ref{sec:intro}, the gripper moves toward the cup and then shifts sideways.
Suppose that the visual description places the fingers slightly to one side of the cup and the recorded robot states show the gripper moves only a short distance in this window.
The position of the fingers relative to the cup helps the selector interpret this small movement as an adjustment for the grasp.
To judge \textit{whether the observations kept at each speedup factor capture this adjustment}, the selector compares how far the gripper travels along its recorded path between successive observations.
If the higher factor saves only one additional step but skips the observation at the change in direction, the selector may choose the lower factor to keep that observation in the training data.
During steady transport through open space, an intermediate observation may add little detail, making the higher factor more appropriate.

\smallskip
\Needspace{3\baselineskip}
\noindent\textbf{Step reduction and data changes.}
To favor acceleration, we instruct the selector to \textit{start with the highest candidate factor}, which is 2.0 in our experiments.
It considers 1.5 when the additional observations or finer commands matter for the required movement, and then considers 1.0 if it preserves further relevant detail.
When two candidates have similar support, the selector prefers the one that removes more steps.
If both candidates group the source steps in the same way, it selects the lower factor.
Appendix~\ref{app:selection_specification} describes when the selector may consider a lower factor.

\smallskip
\noindent\textbf{Full-episode context.}
The purpose of a local adjustment may become clear \textit{from the movement that follows in a later window}.
In the cup example, the sideways adjustment may occur at the end of one window.
If the next window shows the fingers on either side of the cup when the closing command occurs, the selector can interpret the earlier adjustment as preparation for the grasp.
The selector considers the ordered windows together and returns one factor for each window in a single response, without a fixed compression target for the episode.
We accept the profile for retiming only if it assigns one allowed factor to each window in order and every reference in $E$ points to a field in the selector input.

\subsection{Demonstration Retiming}
\label{sec:retiming}

Algorithm~\ref{alg:offline_acceleration} summarizes how \ours{} constructs the reasoning inputs, selects a speedup profile for the episode, and retimes the demonstration.
In the final step, we apply the selected profile with the same operator $\Phi$ and settings used to compute the candidate retiming effects.
Within each window $w$, $\Phi$ groups consecutive source steps into intervals according to the selected factor $r_w$.
The operator starts a new interval at each gripper-command transition and limits compression during measured finger motion.
Each resulting interval $C_j=[s_j,e_j)$ covers source steps $s_j$ through $e_j-1$ and contributes one training pair.

\begin{algorithm}[t]
  \caption{Offline Acceleration with \ours{}}
  \label{alg:offline_acceleration}
  \small
  \begin{algorithmic}[1]
    \algrenewcommand\algorithmicrequire{\textbf{Input:}}
    \algrenewcommand\algorithmicensure{\textbf{Output:}}
    \Require Demonstration $\tau$, task instruction $\ell$, candidate factors $\mathcal R$, retiming operator $\Phi$ with fixed settings, pretrained selector $f_{\mathrm{LLM}}$, selection instructions $\mathcal I$
    \Ensure Profile $\mathbf r$, explanations $E$, accelerated demonstration $\widetilde\tau$
    \Statex \textbf{/* Speedup reasoning inputs */}
    \State Partition $\tau$ into fixed, ordered windows $\mathcal W_{1:W}$
    \For{$w=1,\ldots,W$}
      \State $v_w\gets\operatorname{VisualContext}(\tau,\ell,\mathcal W_w)$
      \State $m_w\gets\operatorname{OriginalMotion}(\tau,\mathcal W_w)$
      \State Compute candidate retiming effects $e_w(r)$ for each $r\in\mathcal R$ using $\Phi$
      \State $x_w\gets(v_w,m_w,\{e_w(r)\}_{r\in\mathcal R})$
    \EndFor
    \Statex \textbf{/* Speedup profile reasoning */}
    \State $(\mathbf r,E)\gets f_{\mathrm{LLM}}(\ell,x_1,\ldots,x_W;\mathcal I)$

    \State Check that $\mathbf r$ lists one allowed factor per window in order
    \State Accept $\mathbf r$ only if this check passes and all fields cited in $E$ exist in the input
    \Statex \textbf{/* Demonstration retiming */}
    \State $\widetilde\tau\gets\Phi(\tau,\mathbf r)$
    \State \Return $\mathbf r,E,\widetilde\tau$
  \end{algorithmic}
\end{algorithm}

\smallskip
For example, suppose $\Phi$ groups steps 4 and 5 of a LIBERO~\cite{liu2023libero} demonstration.
The observation and robot state at step 4 form the input.
To form the action target, we add the two arm delta commands, clip the sum, and use the gripper command from step 5.
For interval $j$ with first step $s_j$ and last step $e_j-1$, the resulting observation and commands are expressed as follows:
\begin{equation}
  \begin{aligned}
    \widetilde{o}_j &= o_{s_j}, \qquad
    \widetilde{g}_j = g_{e_j-1},\\
    \widetilde{\mathbf{u}}_j &=
      \operatorname{clip}_{c}
      \left(\sum_{t=s_j}^{e_j-1}\mathbf{u}_t\right),
  \end{aligned}
  \label{eq:aggregation}
\end{equation}
where $\mathbf{u}_t$ and $g_t$ denote the recorded arm delta and gripper commands, respectively, and $c$ is the command limit.
Appendix~\ref{app:platform_rules} describes the grouping rules shared by all three platforms and the command transformations used for RoboCasa-Kitchen and Franka.

\section{Experiments}
\label{sec:evaluation}

Our experiments address the following two questions.
(1) How do policies fine-tuned on demonstrations accelerated by \ours{} compare to baseline policies in task success rate and completion time?
(2) How do speedup decisions, the resulting training data, and policy performance change when we remove either visual context or candidate retiming effects from the full method?
Sections~\ref{sec:policy_performance} and~\ref{sec:ablation} examine these questions, respectively.

\Needspace{5\baselineskip}
\subsection{Experimental Setup}
\label{sec:experimental_setup}

\smallskip
\Needspace{3\baselineskip}
\noindent\textbf{Simulation tasks.}
We use two representative simulation benchmarks for robot manipulation.
LIBERO~\cite{liu2023libero} tests manipulation under different object arrangements and task goals, while RoboCasa-Kitchen~\cite{nasiriany2024robocasa} models everyday tasks in kitchen environments.
For LIBERO, we use 1,693 demonstrations from 40 tasks in the Spatial, Object, Goal, and Long suites.
For RoboCasa-Kitchen, we use 300 demonstrations for each of 24 tasks, totaling 7,200 demonstrations.
All methods use the same initial-state sampling and 50 evaluation trials per task, totaling 2,000 trials on LIBERO and 1,200 on RoboCasa.

\begin{figure}[!t]
  \centering
  \captionsetup[subfigure]{labelformat=parens,labelsep=none}
  \begin{subfigure}[t]{0.58\linewidth}
    \centering
    \includegraphics[width=\linewidth]{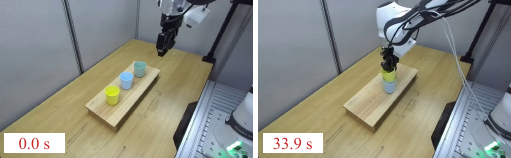}
    \caption{}
    \label{fig:real-task-stack}
  \end{subfigure}
  \par\smallskip
  \begin{subfigure}[t]{0.58\linewidth}
    \centering
    \includegraphics[width=\linewidth]{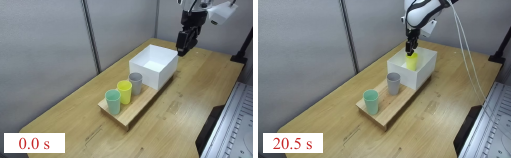}
    \caption{}
    \label{fig:real-task-basket}
  \end{subfigure}
  \par\smallskip
  \begin{subfigure}[t]{0.58\linewidth}
    \centering
    \includegraphics[width=\linewidth]{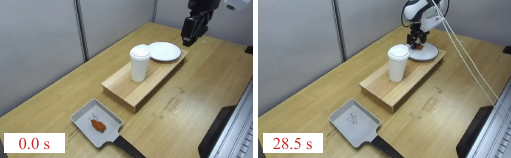}
    \caption{}
    \label{fig:real-task-chicken}
  \end{subfigure}
  \caption{Demonstrations on Franka Research 3 for (a) Stack Cup, (b) Middle Cup to Basket, and (c) Chicken to Plate with a cup obstacle.}
  \label{fig:real-tasks}
\end{figure}

\smallskip
\Needspace{3\baselineskip}
\noindent\textbf{Real-world tasks.}
As shown in Fig.~\ref{fig:real-tasks}, we designed three real-world manipulation tasks for the Franka Research 3 arm.
In Stack Cup, the robot performs two consecutive stacking operations.
In Middle Cup to Basket, it selects the middle cup and places it in the basket.
In Chicken to Plate, it transfers chicken from a pan to a plate with a cup serving as an obstacle.
We collect 40 teleoperated demonstrations for each task and conduct 32 evaluation trials per method.
Appendix~\ref{app:real_protocol} lists the task success criteria and evaluation settings for Franka.

\smallskip
\noindent\textbf{Implementation details.}
\ours{} uses Qwen3.8-27B~\cite{qwen3.8} for speedup profile reasoning and Qwen3-VL-32B-Instruct-FP8~\cite{bai2025qwen3} for visual context extraction.
For policy fine-tuning, we initialize all policies from the same pretrained GR00T N1.6~\cite{gr00tn1_2025} checkpoint.
In simulation, we fine-tune one multitask policy per benchmark under each training condition, using 20,000 optimization steps on LIBERO and 60,000 on RoboCasa.
For each real-world task, we fine-tune a separate policy under each condition for 30,000 optimization steps.
We use the same optimization and inference settings for all conditions within each benchmark.
Appendix~\ref{app:simulation_protocol} lists the simulation training and rollout settings.

\smallskip
\noindent\textbf{Baselines.}
We refer to the GR00T N1.6 policy fine-tuned on the unaccelerated demonstrations as Original.
We evaluate three acceleration baselines: Uniform acceleration, DemoSpeedup~\cite{guo2025demospeedup}, and VOLT~\cite{sanchez2026volt}.
Uniform acceleration applies our retimer with a fixed factor of 2.0 in simulation and a more conservative factor of 1.5 on the physical robot.
We adapt DemoSpeedup and VOLT to the GR00T action interface using their published segmentation criteria and acceleration rules.
Appendix~\ref{app:volt_adaptation} details their segment-level acceleration rates and action conversion procedures in simulation.

\smallskip
\Needspace{3\baselineskip}
\noindent\textbf{Metrics.}
We measure success rate (SR) over all trials and mean completion time (Time) over successful trials.
For Time, we count control steps in simulation and measure elapsed time in seconds on the Franka robot, including policy inference time.
Since Time excludes failed trials, we also use effective speedup (ES)~\cite{jeon2026pacevla} as an efficiency metric in simulation, defined as follows:
\begin{equation*}
  \mathrm{ES} = \frac{T_{\mathrm{eff,orig}}}{T_{\mathrm{eff}}},
\end{equation*}
where $T_{\mathrm{eff}}$ denotes the number of control steps per success for the evaluated policy and $T_{\mathrm{eff,orig}}$ denotes the corresponding value for Original.
For each policy, we count the steps to first success for each successful trial and the task-specific horizon for each failed trial, then divide the total by the number of successes (Appendix~\ref{app:metric_details}).

\begin{table}[!t]
  \centering
  \captionsetup{skip=8pt}
  \caption{Policy performance on four LIBERO suites with GR00T N1.6.}
  \label{tab:libero_suites}
  \fontsize{8.5}{10}\selectfont
  \renewcommand{\arraystretch}{0.9}
  \setlength{\tabcolsep}{0pt}
  \newcommand{\suiteheads}{SR$\uparrow$ & Time$\downarrow$ & ES$\uparrow$}
  \begin{tabular*}{\textwidth}{@{}l*{5}{@{\extracolsep{6pt plus 1.5fill}}c@{\extracolsep{2pt plus 1fill}}cc}@{}}
    \toprule
    & \multicolumn{3}{c}{Spatial}
    & \multicolumn{3}{c}{Object}
    & \multicolumn{3}{c}{Goal}
    & \multicolumn{3}{c}{Long}
    & \multicolumn{3}{c}{Overall} \\
    \cmidrule(lr){2-4}\cmidrule(lr){5-7}\cmidrule(lr){8-10}\cmidrule(lr){11-13}\cmidrule(lr){14-16}
    Method & \suiteheads & \suiteheads & \suiteheads & \suiteheads & \suiteheads \\
    \midrule
    GR00T N1.6 (Original)~\cite{gr00tn1_2025}
    & 98.4 & 100.8 & 1.000
    & 99.4 & 136.2 & 1.000
    & 97.4 & 108.2 & 1.000
    & 94.6 & 262.8 & 1.000
    & 97.45 & 151.0 & 1.000 \\
    \midrule
    \quad + Uniform acceleration
    & 75.0 & 66.4 & 0.747
    & 96.6 & 81.6 & 1.507
    & 82.8 & 66.8 & 0.900
    & 83.8 & 160.6 & \textbf{1.120}
    & 84.55 & 94.2 & 1.053 \\
    \quad + DemoSpeedup~\cite{guo2025demospeedup}
    & 78.0 & 65.0 & 0.822
    & 93.6 & 83.3 & 1.346
    & 84.8 & 72.5 & 0.920
    & 73.2 & 177.4 & 0.795
    & 82.40 & 97.1 & 0.932 \\
    \quad + VOLT~\cite{sanchez2026volt}
    & 68.6 & 60.6 & 0.647
    & 95.4 & 71.6 & \textbf{1.620}
    & 84.2 & 68.4 & 0.932
    & 67.6 & 160.0 & 0.715
    & 78.95 & 87.3 & 0.889 \\
    \quad + \ours{} (Ours)
    & 96.0 & 67.2 & \textbf{1.366}
    & 95.0 & 86.7 & 1.360
    & 94.2 & 69.4 & \textbf{1.322}
    & 84.6 & 168.6 & 1.111
    & 92.45 & 96.0 & \textbf{1.257} \\
    \bottomrule
  \end{tabular*}
\end{table}

\begin{table}[!t]
  \centering
  \captionsetup{skip=8pt}
  \caption{Policy performance on RoboCasa-Kitchen with GR00T N1.6.}
  \label{tab:robocasa_results}
  \fontsize{8.5}{10}\selectfont
     \renewcommand{\arraystretch}{0.9}
  \setlength{\tabcolsep}{0pt}
  \newcommand{\suiteheads}{SR$\uparrow$ & Time$\downarrow$ & ES$\uparrow$}
  \begin{tabular*}{\textwidth}{@{}l*{3}{@{\extracolsep{8pt plus 1.5fill}}c@{\extracolsep{4pt plus 1fill}}cc}@{}}
    \toprule
    & \multicolumn{3}{c}{Pick-and-Place}
    & \multicolumn{3}{c}{Other (16 Tasks)}
    & \multicolumn{3}{c}{Overall} \\
    \cmidrule(lr){2-4}\cmidrule(lr){5-7}\cmidrule(lr){8-10}
    Method & \suiteheads & \suiteheads & \suiteheads \\
    \midrule
    GR00T N1.6 (Original)~\cite{gr00tn1_2025}
    & 48.0 & 376.5 & \textbf{1.000}
    & 78.1 & 288.9 & 1.000
    & 68.1 & 309.5 & 1.000 \\
    \midrule
    \quad + Uniform acceleration
    & 35.2 & 219.1 & 0.709
    & 73.6 & 184.4 & 1.047
    & 60.8 & 191.1 & 0.931 \\
    \quad + DemoSpeedup~\cite{guo2025demospeedup}
    & 29.2 & 240.7 & 0.548
    & 72.1 & 203.8 & 0.963
    & 57.8 & 210.0 & 0.828 \\
    \quad + VOLT~\cite{sanchez2026volt}
    & 31.5 & 228.7 & 0.607
    & 69.2 & 159.5 & 0.941
    & 56.7 & 172.4 & 0.829 \\
    \quad + \ours{} (Ours)
    & 41.2 & 245.7 & 0.873
    & 76.0 & 204.7 & \textbf{1.092}
    & 64.4 & 213.5 & \textbf{1.015} \\
    \bottomrule
  \end{tabular*}
\end{table}

\subsection{Main Results}
\label{sec:policy_performance}
We first compare policies trained on original and accelerated demonstrations in simulation, then examine their success rates and completion times on Franka.

\subsubsection{Simulation Benchmarks}
\label{sec:simulation_results}

To evaluate the effect of demonstration acceleration on policy performance, we fine-tune a separate GR00T N1.6 policy for each training condition using demonstrations from all 40 LIBERO tasks.
As shown in Table~\ref{tab:libero_suites}, \ours{} achieves the highest overall ES of 1.257 and the highest success rate among policies trained on accelerated demonstrations (92.45\%).
Compared to Original, \ours{} reduces the completion time of successful trials by 36.4\% (from 151.0 to 96.0), while its success rate decreases by 5.13\% (from 97.45\% to 92.45\%).

To jointly assess success rate and completion time, we examine the ratio $\mathrm{SR}/\mathrm{Time}$.
All four acceleration methods improve $\mathrm{SR}/\mathrm{Time}$ compared to Original, with increases ranging from 31.5\% for DemoSpeedup (from 0.645 to 0.849) to 49.2\% for \ours{} (from 0.645 to 0.963).
Although VOLT achieves the shortest completion time of 87.3, it also has the lowest SR of 78.95\%.
When each failed trial contributes its task-specific horizon, VOLT has the lowest overall ES of 0.889.
\ours{} is the only method with ES above 1 on both Spatial and Goal, reaching 1.366 and 1.322, respectively.
Its success rate exceeds 94\% on both suites, compared to less than 85\% for each of the other acceleration baselines.
The largest decrease in success rate for \ours{} occurs on Long, where all accelerated policies fall below Original.
In multistage tasks, a positioning error during one movement may affect subsequent actions.
\ours{} achieves the highest success rate among accelerated policies on Long at 84.6\%.

We next examine whether demonstration acceleration improves execution efficiency on RoboCasa-Kitchen, where Original success rate is 68.1\% compared to 97.45\% on LIBERO.
As shown in Table~\ref{tab:robocasa_results}, all four acceleration methods shorten successful trials compared to Original, but only \ours{} achieves an overall ES above 1 (1.015).
Among the accelerated policies, \ours{} achieves the highest success rate in both task groups, averaging 41.2\% on the eight pick-and-place tasks and 76.0\% on the other 16 tasks.
On the pick-and-place tasks, successful trials with \ours{} take 34.7\% fewer steps on average than with Original, but the success rate falls from 48.0\% to 41.2\%, bringing ES down to 0.873.
On the other tasks, the success rate decreases only from 78.1\% to 76.0\%, so a reduction of 29.1\% (from 288.9 to 204.7) in the mean number of steps in successful trials is sufficient to raise ES to 1.092.

\subsubsection{Real-World Robot Evaluation}
\label{sec:real_results}
\begin{table}[H]
  \centering
  \captionsetup{skip=8pt}
  \caption{Real-world policy performance on three manipulation tasks with Franka Research 3.
  Time denotes mean completion time in seconds among successful trials.}
  \label{tab:real_results}
  \fontsize{8.5}{10}\selectfont
   \renewcommand{\arraystretch}{0.9}
  \setlength{\tabcolsep}{0pt}
  \newcommand{\taskheads}{\makebox[.09\linewidth]{SR$\uparrow$} & \makebox[.09\linewidth]{Time (s)$\downarrow$}}
  \begin{tabular*}{\textwidth}{@{}p{.25\linewidth}*{3}{@{\extracolsep{8pt plus 1.5fill}}c@{\extracolsep{4pt plus 1fill}}c}@{}}
    \toprule
    & \multicolumn{2}{c}{Stack Cup}
    & \multicolumn{2}{c}{Middle Cup to Basket}
    & \multicolumn{2}{c}{Chicken to Plate} \\
    \cmidrule(lr){2-3}\cmidrule(lr){4-5}\cmidrule(lr){6-7}
    Method & \taskheads & \taskheads & \taskheads \\
    \midrule
    GR00T N1.6 (Original)~\cite{gr00tn1_2025}
    & 71.9 & 41.6
    & 81.3 & 26.8
    & 62.5 & 23.8 \\
    \midrule
    \quad + Uniform acceleration
    & 62.5 & 26.8
    & 68.8 & 17.2
    & 59.4 & 17.7 \\
    \quad + DemoSpeedup~\cite{guo2025demospeedup}
    & 53.1 & 32.0
    & 71.9 & 20.7
    & 56.3 & 20.9 \\
    \quad + VOLT~\cite{sanchez2026volt}
    & 59.4 & 31.4
    & 62.5 & 20.7
    & 53.1 & 20.3 \\
    \quad + \ours{} (Ours)
    & 68.8 & 21.4
    & 75.0 & 14.3
    & 59.4 & 14.8 \\
    \bottomrule
  \end{tabular*}
\end{table}

The three Franka tasks combine object transport with alignment or placement, allowing us to examine how demonstration acceleration affects task success and completion time on a physical robot.
As shown in Table~\ref{tab:real_results}, \ours{} completes successful trials faster on average than Uniform acceleration, DemoSpeedup, and VOLT on all three tasks while matching or exceeding their success rates.
On Chicken to Plate, \ours{} and Uniform acceleration each succeed in 19 of 32 trials, with mean completion times of 14.8 and 17.7 seconds, respectively.
Compared to Original, \ours{} reduces the mean completion time of successful trials on Stack Cup by 48.6\% (from 41.6 to 21.4 seconds), with one fewer success in 32 trials.
One possible explanation is that \ours{} can assign a higher speedup factor during transport and a lower factor when alignment requires finer control.
When two candidate commands reach the same controller limit, the offline selector can compare their peaks before clipping and use the visible clearance to assess whether the higher factor is suitable for alignment.

\subsection{Ablation Study}
\label{sec:ablation}

To examine how candidate retiming effects and visual context inform speedup decisions, we remove one input at a time on LIBERO (Table~\ref{tab:ablation_results}).
Both variants use the task instruction and original motion measurements (Appendix~\ref{app:compared_conditions}).
We use the same LLM, candidate factors, retiming operator, and training and evaluation settings for all three conditions.

\begin{table}[H]
  \centering
  \caption{Ablation study of speedup reasoning inputs on LIBERO.
  Compression is the ratio of source to retimed demonstration steps.}
  \label{tab:ablation_results}
  \fontsize{8.5}{10}\selectfont
  \renewcommand{\arraystretch}{0.9}
  \setlength{\tabcolsep}{0pt}
  \newcommand{\ablationheads}{%
    \makebox[.075\linewidth]{SR$\uparrow$} &
    \makebox[.075\linewidth]{Time$\downarrow$} &
    \makebox[.075\linewidth]{ES$\uparrow$}}
  \begin{tabular*}{\textwidth}{@{}l@{\hspace{10pt}}c*{3}{@{\extracolsep{\fill}}c@{\extracolsep{0pt}}cc}@{}}
    \toprule
    & & \multicolumn{3}{c}{Short (Spatial, Object, Goal)}
    & \multicolumn{3}{c}{Long}
    & \multicolumn{3}{c}{Overall} \\
    \cmidrule(lr){3-5}\cmidrule(lr){6-8}\cmidrule(lr){9-11}
    Variant & Compression & \ablationheads & \ablationheads & \ablationheads \\
    \midrule
    \ours{} (Ours) & 1.718
    & 95.07 & 74.4 & \textbf{1.351}
    & 84.6 & 168.6 & \textbf{1.111}
    & 92.45 & 96.0 & \textbf{1.257} \\
    w/o candidate effects & 1.682
    & 92.60 & 77.6 & 1.201
    & 83.8 & 179.8 & 1.044
    & 90.40 & 101.2 & 1.142 \\
    w/o visual context & 1.766
    & 94.20 & 72.2 & 1.347
    & 82.6 & 161.5 & 1.079
    & 91.30 & 92.4 & 1.243 \\
    \bottomrule
  \end{tabular*}
\end{table}

\smallskip
\Needspace{3\baselineskip}
\noindent\textbf{Impact of candidate retiming effects.}
Removing candidate retiming effects lowers compression from 1.718 to 1.682.
Despite the additional demonstration steps available for fine-tuning, the overall success rate falls from 92.45\% to 90.40\%.
Successful trials also take more control steps on both Short and Long, and overall ES falls from 1.257 to 1.142.
These results suggest that even with the recorded scene and motion available, the selector benefits from comparing how the candidate factors would change the training data.

\smallskip
\noindent\textbf{Impact of visual context.}
With visual context, the selector compresses the demonstrations less, and the resulting policy achieves a higher success rate on both Short and Long.
On Long, for example, the success rate increases from 82.6\% to 84.6\%, while the mean number of control steps in successful trials increases from 161.5 to 168.6.
The higher success rate and longer successful trials result in similar overall ES values with and without visual context (1.257 and 1.243, respectively).

\begin{figure}[!t]
  \centering
  \captionsetup[subfigure]{labelformat=parens,labelsep=none}
  \begin{subfigure}[t]{0.49\linewidth}
    \centering
    \includegraphics[trim=0 10pt 0 2pt,clip,width=1.0\linewidth]{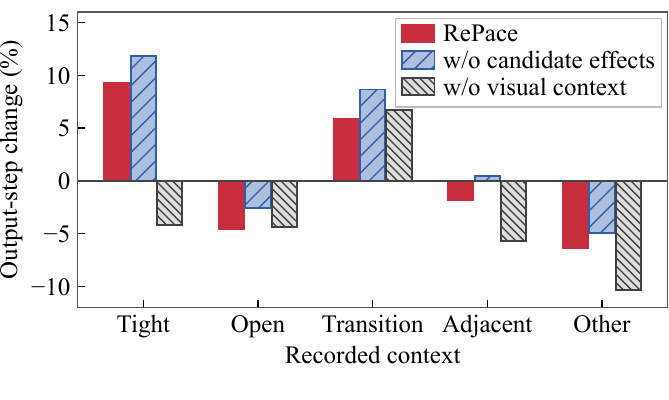}
    \caption{}
    \label{fig:decision_context}
  \end{subfigure}\hfill%
  \begin{subfigure}[t]{0.49\linewidth}
    \centering
    \includegraphics[trim=0 10pt 0 2pt,clip,width=1.0\linewidth]{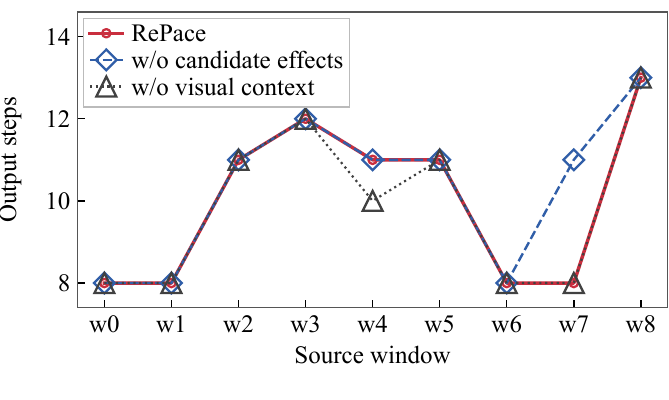}
    \caption{}
    \label{fig:decision_episode}
  \end{subfigure}
  \caption{Output steps in LIBERO demonstrations by (a) recorded context, compared to uniform acceleration at similar overall compression, and (b) source window within one demonstration.}
  \label{fig:decision_data}
\end{figure}

\noindent\textbf{Analysis of speedup decisions.}
To provide an intuitive understanding of how visual context and candidate retiming effects guide speedup selection, we compare the number of demonstration steps kept by the full method and its two variants.
As shown in Fig.~\ref{fig:decision_data}\subref{fig:decision_context}, the full method and the variant without visual context show similar reductions in demonstration steps when the robot has more room to move.
When space is limited (Tight), however, the full method keeps more steps than the variant without visual context.

Figure~\ref{fig:decision_data}\subref{fig:decision_episode} shows the output steps for a demonstration of placing cream cheese in a bowl.
\ours{} and its two ablation variants produce the same number of output steps in seven of the nine windows.
In w4, the variant without visual context produces 10 steps, while the other two produce 11.
In w7, increasing the factor from 1.5 to 2.0 reduces the output from 11 steps to 8 and increases the maximum distance the gripper travels between successive observations (Appendix~\ref{app:decision_data_analysis}).
Without candidate retiming effects, the selector chooses 1.5 and explains its choice in terms of alignment near the bowl.
With the candidate measurements available, \ours{} chooses 2.0 and explains that the gripper has room to move along the path.

\Needspace{7\baselineskip}
\subsection{Discussion and Limitations}
\label{sec:discussion}
Before VLA fine-tuning, \ours{} uses the recorded scene and motion to judge how much to accelerate each part of a demonstration.
With fixed boundaries, all candidates start from the same observations and commands, so their effects can be compared directly.
When a window includes both transport and alignment, however, one factor applies to both movements.
Adaptive boundaries could follow such transitions, allowing the model to choose a separate factor for each movement.

\section{Conclusion}
\label{sec:conclusion}
We have presented \ours{}, which selects demonstration speedup profiles before policy fine-tuning without an additional speed-selection module at deployment.
For each movement, \ours{} uses the recorded scene and motion to judge whether the control steps saved by a higher speedup factor justify the resulting changes to observations and action targets.
On LIBERO, the policy trained with \ours{} requires 36.4\% fewer control steps on average in successful trials than Original and achieves an overall effective speedup of 1.257.
On the Franka Stack Cup task, \ours{} reduces the mean completion time of successful trials by 48.6\% compared to Original, with one fewer success in 32 trials.
In the ablation study, the full method achieves a higher success rate and a shorter mean completion time of successful trials than the variant without candidate retiming effects.
In future work, we will investigate how to select window boundaries together with speedup factors.

\section*{LLM Usage Statement}

We used ChatGPT and Claude to help draft and revise the text, check related work and citations, and refine code.
We reviewed the results and made the final scientific judgments.

{\small
\setlength{\bibsep}{2pt plus 1pt minus 1pt}
\bibliographystyle{ieeenat_fullname}
\bibliography{main}}
\clearpage
\appendix

\section{Reasoning Inputs and Selection}
\label{app:cue_definitions}
This section describes the three inputs for each source window and how \ours{} uses them to select a speedup profile for the episode.
Each source window covers steps $s_w$ through $e_w-1$.
We merge a final fragment shorter than eight steps with the preceding window and mark missing or invalid measurements as unknown.

\subsection{Visual Context}
\label{app:source_cues}
\label{app:visual_context}
The visual description records object arrangement, obstacles, route clearance, and visibility (Table~\ref{tab:visual_extraction_settings}).
Route clearance can be open, tight, or unknown.
The template lists clear, partial, and occluded for visibility.
If the response provides none of these labels, we record visibility as unknown.
The instruction templates identify the objects and destinations, using \emph{fixture part} as the destination for fixture operations.
Context frames come from the same episode and carry camera and source-frame identifiers.
We omit arrangement descriptions containing designated contact, grasp, support, or completion terms, while keeping the other fields and their frame references.

\subsection{Original Motion}
\label{app:original_motion}
We record the initial, final, minimum, and maximum signed finger aperture using the definitions and units in Table~\ref{tab:numerical_input_mappings}.
For each window of $n_w=e_w-s_w$ observations, we measure end-effector displacement, vertical change, and path length from world-frame positions $p_t$ in meters.
We compute mean and maximum speeds from step lengths divided by the sampling interval $\Delta t$ in seconds.
Let $R_t$ denote the recorded end-effector rotation matrix at step $t$, and let $d_R(R_t,R_{t+1})$ denote the angle of the relative rotation $R_t^\top R_{t+1}$.
We compute angular path length and maximum angular speed as
\begin{equation}
  \begin{aligned}
    L_w^R &= \sum_{t=s_w}^{e_w-2} d_R(R_t,R_{t+1}),\\
    \omega_w^{\max} &=
      \max_{s_w\leq t<e_w-1} d_R(R_t,R_{t+1})/\Delta t.
  \end{aligned}
  \label{eq:source_rotation}
\end{equation}
We express these quantities in radians and radians per second, respectively.

We record the source frame and direction of each gripper-command transition.
For each window, we count its transitions and measure the distances to the nearest preceding and following transitions in both frames and windows.

\subsection{Candidate Retiming Effects}
\label{app:candidate_cues}
\label{app:candidate_effects}
Let $n_w(r)$ denote the number of output steps in window $w$ after $\Phi$ applies the gripper rules at factor $r$.
Relative to the next lower factor $r^{-}$, candidate $r$ saves $n_w(r^{-})-n_w(r)$ additional steps.
For the lowest factor, the comparison uses the original window length $n_w$.
To measure peak requested motion in m/s and rad/s, we convert the maximum stored command norms using Table~\ref{tab:numerical_input_mappings} and divide by the control period.
The selector receives post-clipping peaks and the number of output steps affected by clipping.
For Franka, it also receives pre-clipping peaks when clipping occurs.

For successive selected observations at source steps $i<j$, with $i$ in window $w$, we compute the translational and angular path lengths along the recorded motion as
\begin{equation}
  \begin{aligned}
    G_p(i,j) &= \sum_{t=i}^{j-1}\|p_{t+1}-p_t\|_2,\\
    G_R(i,j) &= \sum_{t=i}^{j-1}d_R(R_t,R_{t+1}).
  \end{aligned}
  \label{eq:retained_motion_spans}
\end{equation}
For each candidate, we record the maximum distance traveled (m), the maximum angular path length (rad), and the source interval for each maximum.
A path can extend to the first selected observation in the next window. The final path ends at the last recorded state.
Gripper mappings record the source frame, transition type, and corresponding output step.
We mark two candidates as output-equivalent only if their ordered source-interval partitions match after finger-motion handling and transition splits.
Candidates with the same output count can have different partitions.

\subsection{Selection Rules}
\label{app:selection_specification}
The selector first identifies the required spatial relation, then examines the recorded motion and candidate effects.
It may consider a lower factor when at least one of the following conditions holds.
(i) Motion in the current or a neighboring window forms a required spatial relation between the held or targeted object and its destination or a fixture part, as in grasp alignment, placement, insertion, or manipulation of a handle or fixture.
(ii) The visual context indicates tight clearance along the path of transport or rotation.
(iii) Motion before or after a gripper command in the same window meets condition (i).
The explanation identifies the applicable condition.
If none applies, the instructions select 2.0, subject to the identical-output rule in Section~\ref{sec:reasoning}.

When a condition holds, the selector explains why the additional observations or finer commands matter for the required movement.
A larger command peak, a longer path between observations, a gripper transition, an object in hand, contact, or missing information alone cannot justify a lower factor.

\newenvironment{instructionquote}{%
  \begingroup
  \setlength{\leftmargini}{8pt}%
  \begin{quote}
  \small\raggedright
  \fontencoding{T1}\selectfont
  \setlength{\parskip}{4pt}%
  \hrule height 0.4pt\nobreak\smallskip
}{%
  \par\nobreak\smallskip\hrule height 0.4pt
  \end{quote}
  \endgroup
}

\noindent\begin{minipage}{\textwidth}
  \centering
  \captionof{table}{Settings for extracting visual context from LIBERO, RoboCasa-Kitchen, and Franka demonstrations.}
  \label{tab:visual_extraction_settings}
  \small
  \setlength{\tabcolsep}{4pt}
  \renewcommand{\arraystretch}{1.12}
  \begin{tabular}{@{}p{0.19\textwidth}p{0.19\textwidth}p{0.26\textwidth}p{0.28\textwidth}@{}}
    \toprule
    Setting & LIBERO & RoboCasa-Kitchen & Franka \\
    \midrule
    Camera views & External and wrist & Left external, right external, and wrist &
    Wrist and external, left image of each stereo pair \\
    Instruction parser & \raggedright Object, destination, and fixture templates &
    \raggedright Pick-and-place and fixture templates &
    \raggedright Stacking and pick-and-place templates \tabularnewline
    Image size & $256\times256$ & $256\times256$ & $576\times320$, resized from $1280\times720$ \\
    Source recording rate & 20 Hz & 20 Hz & 10 Hz \\
    \midrule
    Visual model & \multicolumn{3}{c}{Qwen3-VL-32B-Instruct-FP8} \\
    Nominal source window & \multicolumn{3}{c}{16 steps} \\
    Context & \multicolumn{3}{c}{Up to 8 source steps before and after the window} \\
    Frame sampling & \multicolumn{3}{c}{Stride 2, with the final context frame included} \\
    Temperature & \multicolumn{3}{c}{0} \\
    Maximum output tokens & \multicolumn{3}{c}{220} \\
    \bottomrule
  \end{tabular}
\end{minipage}
\vspace{12pt}
\subsection{Instruction Templates}
\label{app:instruction_templates}
We reproduce the visual context extraction template and an excerpt of the speedup profile selection instructions.
The complete selection instructions also define the input fields, platform context, and JSON response format.

\par\medskip
\noindent\textbf{Visual context extraction.}
Named placeholders are filled from each episode and window.
The template uses 20 frames per second for simulation and 10 for Franka, as specified in Table~\ref{tab:visual_extraction_settings}.
\Needspace{8\baselineskip}
\begin{instructionquote}
You are given synchronized frames of one robot manipulation demonstration from \{n\_cams\} camera(s): \{camera\_list\}. The instruction given to the robot was '\{instruction\}'; use it only to know which object and destination are meant.
Every image is labelled with its absolute frame index (20 frames per second). Frames shown: \{frame\_list\}. The TARGET window is frames \{t\_start\} to \{t\_end\_incl\}; the other frames are context before and after it.
\par
Describe only what is VISIBLE in the TARGET window, as a JSON object:
\{\textquotedbl{}target\_goal\_arrangement\textquotedbl{}: \textless{}at most 25 words: where the \{target\} and the \{goal\} are on screen relative to each other and to the gripper, e.g. \textquotedbl{}bowl on the table left of the plate, gripper above the bowl\textquotedbl{}\textgreater{},
\par
\textquotedbl{}obstacles\textquotedbl{}: [\textless{}names of nearby objects or fixture parts that constrain the motion; empty list if none\textgreater{}],
\par
\textquotedbl{}route\_clearance\textquotedbl{}: \textquotedbl{}open\textquotedbl{} \textbar{} \textquotedbl{}tight\textquotedbl{} \textbar{} \textquotedbl{}unknown\textquotedbl{} (free space around the gripper or the carried object along its motion),
\par
\textquotedbl{}visibility\textquotedbl{}: \textquotedbl{}clear\textquotedbl{} \textbar{} \textquotedbl{}partial\textquotedbl{} \textbar{} \textquotedbl{}occluded\textquotedbl{},
\par
\textquotedbl{}evidence\_frames\textquotedbl{}: [\textless{}the shown frame indices that support the description\textgreater{}]\}
\par
Rules: write \textquotedbl{}unknown\textquotedbl{} (or an empty description) whenever something is not visible, occluded, too small or ambiguous. Do not state whether the object is touched, grasped, held, lifted, supported, placed or released, and do not say whether the task is done or in progress; describe placement as seen. Do not give distances, sizes or tolerances in numbers. Do not use the instruction to guess what should be happening. Output the JSON object only.
\end{instructionquote}

\noindent\textbf{Speedup profile selection.}
The following core instructions are shared by LIBERO, RoboCasa-Kitchen, and Franka.
\begin{instructionquote}
Choose one eligible candidate per window from \{1.0, 1.5, 2.0\}. 1.0 keeps the window's recorded timing; its extra\_steps\_saved is 0.
Start every window from 2.0 and state the steps it saves (extra\_steps\_saved); 2.0 is a choice with evidence, not a default without one. Then ask whether there is a reason to keep more of the recorded example by choosing 1.5, and only then whether 1.0 keeps something for that motion that 1.5 still removes.

Judge in this order:
\par
1. the spatial relation the task requires this window and its neighbours to establish (from task\_context and    visible\_geometry: object to destination, object to gripper, fixture part to its handle);
\par
2. the recorded motion that establishes it (measured\_eef\_motion, measured\_eef\_rotation, measured\_gripper,    gripper\_transitions\_source and nearest\_gripper\_command);
\par
3. what each candidate changes in the training data for that motion (output\_steps and extra\_steps\_saved,    output\_command\_peak, largest\_recorded\_motion\_span, gripper\_command\_mapping);
\par
4. the factor.

A lower factor is considered only in these situations, and the evidence names which one applies:
\par
(a) the window or a neighbouring window establishes a required relation between the held or targeted object and     its destination or fixture part: aligning for a grasp, placing or inserting, engaging a handle or fixture;
\par
(b) transport or rotation through tight clearance named in visible\_geometry;
\par
(c) a gripper command whose preceding or following motion in the same window belongs to (a).
Outside these situations choose 2.0. Inside them, the lower factor still needs the comparison below.

A reason for a lower factor connects the required relation and the recorded motion to a candidate-specific change: which retained observations or merged commands of that motion the lower factor keeps. A larger command peak, a larger recorded-motion span, a gripper transition, an object in hand, contact or missing information is not sufficient by itself; connected to the relation and the motion, each of them can be part of the reason.
For the selected factor, whichever it is, relate its step savings and its command or observation changes to the required relation and the recorded motion in the evidence.
When the comparison supports two candidates similarly, choose the one that saves more steps. When candidates produce identical outputs, choose the lower nominal factor.
A larger recorded-motion span means more recorded motion passes between two consecutive retained observations; it is not a tracking error, a contact, a safety margin or a success estimate.

The retimer preserves gripper command transitions and caps compression of the measured finger-motion spans at 1.5.
These rules are already included in the candidate fields. A gripper transition alone does not require further slowing.

Select the full episode profile while considering neighbouring windows and their boundaries.
Do not impose uniform rates, fixed slow buffers or a fixed total compression target.
Do not infer grasp, support or task completion from geometry or commands alone.
Return the existing JSON with brief evidence and valid input references.
\end{instructionquote}

The identical-output rule also applies outside these situations.

\section{Retiming Interfaces}
\label{app:platform_rules}
Table~\ref{tab:numerical_input_mappings} lists the platform-specific action representations and units.
On all three platforms, \ours{} caps the selected speedup factor at 1.5 from each gripper-command transition through the first stable step, up to 40 steps after the transition.
We consider the finger aperture stable at the first step followed by three consecutive aperture changes smaller than $2\times10^{-4}$ in absolute value, using the units in Table~\ref{tab:numerical_input_mappings}.

\begin{table}[t]
  \centering
  \caption{Action representations and unit conversions for original motion and candidate retiming effects.}
  \label{tab:numerical_input_mappings}
  \small
  \setlength{\tabcolsep}{4pt}
  \renewcommand{\arraystretch}{1.12}
  \begin{tabular}{@{}p{0.19\textwidth}p{0.22\textwidth}p{0.27\textwidth}p{0.24\textwidth}@{}}
    \toprule
    Quantity & LIBERO & RoboCasa-Kitchen & Franka \\
    \midrule
    Recorded orientation & Axis-angle &
    World-frame quaternion $(x,y,z,w)$ & Euler angles, extrinsic $xyz$ \\
    Finger aperture & \multicolumn{2}{p{0.49\textwidth}}{Signed difference
    between the two recorded finger positions (m)} &
    One minus the recorded gripper position (unitless) \\
    Arm commands & Translation and rotation deltas &
    Translation and rotation deltas & Normalized linear and angular velocities \\
    Translation / rotation per command unit & 0.05~m / 0.5~rad & 0.05~m / 0.5~rad &
    Estimated 0.0255~m / 0.056~rad per step \\
    Gripper command & $-1$ open, $+1$ close & $0$ open, $1$ close &
    Target from $0$ open to $1$ close; transition at $0.5$ \\
    \ours{} command range & $\pm3$ & $\pm3$ & $\pm1$, not widened \\
    \bottomrule
  \end{tabular}
\end{table}

\Needspace{5\baselineskip}
Within each window, we start a new interval whenever the gripper command changes or the cap ends.
At a factor of 1.5, we divide an interval of $n$ source steps into $\lceil n/1.5\rceil$ contiguous groups by rounding equally spaced boundaries to the nearest step.

\subsection{Simulation}
For LIBERO, Eq.~\ref{eq:aggregation} adds the stored rotation commands component by component without composing the individual rotations.
For RoboCasa, we also sum the base velocity commands and use the final torso target and control mode in each interval.
All three are zero in the 7,200 selected demonstrations. The world-frame end-effector measurements include the recorded base drift.

\subsection{Franka}
\label{app:franka_interface}
For each source interval, we sum the normalized linear and angular velocity commands component by component, clip each sum to $[-1,1]$, and use the final gripper target.
To express the requested motion in physical units, we estimate the command scales by least squares from the 80 Stack Cup and Middle Cup to Basket demonstrations, allowing for a one-step lag between commands and measured motion.

\section{Experimental Details}
\label{app:experimental_parameters}
\subsection{Selector Inference}
For LIBERO, RoboCasa, and Franka, we run the Qwen3.8-27B selector with BF16 weights, temperature 0, top-$p$ 1.0, and thinking disabled.
The context limit is 49,152 tokens, or 65,536 for longer RoboCasa demonstrations, including an 8,192-token output budget.
We do not truncate the input.
Responses follow a JSON schema, with at most one corrective request.
If the selector returns no valid profile, we use the original demonstration.

\subsection{Simulation Training and Evaluation}
\label{app:simulation_protocol}
For simulation, we use NVIDIA's native N1.6.1 release and GR00T-N1.6-3B.
Tables~\ref{tab:n16_training_parameters} and~\ref{tab:n16_rollout_parameters} list the training and evaluation settings.
For all LIBERO tasks, we set the controller dynamics scale to 1 and use the default proportional gain ($k_p = 150$) of the operational space controller.
We exclude the 10 settling steps from completion times and episode limits.
In all simulation conditions, the policy uses four denoising steps, predicts 16 actions, and uses native checkpoint preprocessing for $256\times256$ RGB images.

\begin{table}[t]
  \centering
  \caption{GR00T N1.6 fine-tuning settings on LIBERO and RoboCasa-Kitchen.}
  \label{tab:n16_training_parameters}
  \small
  \setlength{\tabcolsep}{4pt}
  \renewcommand{\arraystretch}{1.08}
  \begin{tabular}{@{}p{0.27\textwidth}p{0.32\textwidth}p{0.36\textwidth}@{}}
    \toprule
    Parameter & LIBERO & RoboCasa-Kitchen \\
    \midrule
    Optimizer updates & 20,000 & 60,000 \\
    Global batch / GPUs & 32 / 2 & 64 / 4 \\
    Warmup updates & 1,000 (5\%) & 3,000 (5\%) \\
    \midrule
    \multicolumn{3}{@{}l}{\emph{Shared settings}} \\
    Microbatch / accumulation & \multicolumn{2}{p{0.68\textwidth}@{}}{16 per GPU / 1} \\
    Optimizer / schedule & \multicolumn{2}{p{0.68\textwidth}@{}}{AdamW / cosine decay} \\
    Learning rate / weight decay & \multicolumn{2}{p{0.68\textwidth}@{}}{$10^{-4}$ / $10^{-5}$} \\
    Frozen modules & \multicolumn{2}{p{0.68\textwidth}@{}}{Vision backbone and lower 12 language-model layers} \\
    Trainable modules & \multicolumn{2}{p{0.68\textwidth}@{}}{Top four language-model layers, projector, and action expert} \\
    Precision / state dropout & \multicolumn{2}{p{0.68\textwidth}@{}}{BF16 / 0} \\
    Color jitter & \multicolumn{2}{p{0.68\textwidth}@{}}{Brightness 0.3, contrast 0.4, saturation 0.5, hue 0.08} \\
    Evaluated checkpoint & \multicolumn{2}{p{0.68\textwidth}@{}}{Final checkpoint} \\
    \bottomrule
  \end{tabular}
\end{table}
\begin{table}[t]
  \centering
  \caption{GR00T N1.6 evaluation settings on LIBERO and RoboCasa-Kitchen.}
  \label{tab:n16_rollout_parameters}
  \small
  \setlength{\tabcolsep}{4pt}
  \renewcommand{\arraystretch}{1.08}
  \begin{tabular}{@{}p{0.27\textwidth}p{0.32\textwidth}p{0.36\textwidth}@{}}
    \toprule
    Parameter & LIBERO & RoboCasa-Kitchen \\
    \midrule
    Trials per condition & 2,000 (500 per suite) & 1,200 (400 PnP, 800 Other) \\
    Policy state / action & Raw axis-angle state / unscaled delta action &
      16-D relative end-effector, gripper, and base state / 12-D action \\
    Action execution & Replan every 5 steps; no temporal action compression &
      Execute all 16 actions per inference call \\
    Command range & $\pm3$ for Original, Uniform, and \ours{};
      $\pm4$ for DemoSpeedup and VOLT &
      $\pm3$ for Original, Uniform, and \ours{};
      $\pm4$ for DemoSpeedup and VOLT \\
    Command scaling & \multicolumn{2}{p{0.68\textwidth}@{}}{0.05~m or 0.5~rad per unit, unchanged from the default controller} \\
    Episode limit & Spatial 220, Object 280, Goal 300, Long 520 action steps &
      1,000 environment steps; failures counted as 1,008 steps at chunk resolution \\
    Completion time & Executed action steps & Executed chunks $\times$ 16 steps \\
    Scene sampling & Shared task and initial-state mapping &
      AI-generated textures; object split A; all kitchens; styles 0 to 8 and 11 \\
    \bottomrule
  \end{tabular}
\end{table}

The 16 RoboCasa tasks in the Other group involve doors, drawers, levers, knobs, buttons, or insertion. The insertion tasks also require cup transport.

\subsection{Metrics and Aggregation}
\label{app:metric_details}
For $M$ simulation trials under a fixed distribution of tasks and initial states, let $s_j\in\{0,1\}$ indicate success in trial $j$.
For a successful trial, $t_j$ is the number of control steps to first success.
For each task, all methods use the same failure cost $H_j$ in executed steps.
When at least one trial succeeds, we compute the number of control steps per success as
\begin{equation}
  T_{\mathrm{eff}} =
  \frac{\displaystyle\sum_{j:\,s_j=1} t_j + \displaystyle\sum_{j:\,s_j=0} H_j}
       {\displaystyle\sum_{j=1}^{M} s_j}.
  \label{eq:evaluation_metrics}
\end{equation}
We assign the full failure cost even when a failed trial ends early.
For RoboCasa, we round the 1,000-step environment limit up to a whole number of 16-step chunks, so $H_j=1{,}008$ (Table~\ref{tab:n16_rollout_parameters}).
We set ES to zero when no trial succeeds and at least one corresponding Original trial succeeds.

Short pools trials from Spatial, Object, and Goal, while Overall pools all trials.
Each comparison uses the corresponding Original trials and task horizons to compute ES as defined in Section~\ref{sec:experimental_setup}.

\subsection{DemoSpeedup and VOLT in Simulation}
\label{app:volt_adaptation}
DemoSpeedup~\cite{guo2025demospeedup} estimates per-step entropy with a Gaussian kernel over overlapping predictions from ten 16-step chunks sampled per observation by each benchmark's Original policy.
After outlier correction, HDBSCAN clusters steps by time and entropy.
Clusters with mean entropy below the episode mean form maintain-speed segments; all remaining steps form speed-up segments.

We implement VOLT~\cite{sanchez2026volt} with the published instructions and omit the unavailable in-context examples.
We run Qwen3-VL-32B-Instruct-FP8 with top-$p$ 0.8, top-$k$ 20, temperature 0.4, and a presence penalty of 0.5.
For LIBERO, we use every frame of the front-camera video at 20~Hz.
For RoboCasa, we subsample the left external camera video from 20 to 10~Hz and map segment timestamps back to source steps.
Both benchmarks use one $256\times256$ video per demonstration.

Following the VOLT comparison with DemoSpeedup, we use 2$\times$ for maintain-speed segments and 4$\times$ for speed-up segments in both methods.
For both, we use trajectories from all temporal offsets, split segments at 16-step window boundaries and gripper-command transitions, and use a command range of $\pm4$ without a finger-motion cap.
We adapt absolute-target skipping to delta commands by summing the deltas in each interval.
\Needspace{3\baselineskip}
We assign the maintain-speed label to overlaps and gaps between different labels and extend the nearest label to uncovered ends.

\subsection{Compared Conditions}
\label{app:compared_conditions}
Uniform and \ours{} both split intervals at gripper-command transitions. Only \ours{} applies the finger-motion cap.
On RoboCasa, Uniform also splits intervals at control-mode changes.
We recompute the normalization statistics for each training dataset.

Both ablations use the task instruction and original motion. We remove the instructions associated with each omitted input.
\emph{Without candidate effects} removes all candidate measurements, including output counts and equivalence markers.
The context rules still apply. When two factors have similar support, the selector favors the higher factor.
\emph{Without visual context} removes VLM cues, duplicated scene descriptions, and the tight-clearance condition in Appendix~\ref{app:selection_specification}.

\subsection{Franka Training and Evaluation}
\label{app:real_protocol}
On Franka, we train with a batch size of 64 (16 samples on each of four GPUs) and evaluate the final checkpoint.
The policy observes external and wrist images, end-effector state, and gripper position.
We encode Euler angles using their sine and cosine.
We use a 95\% image crop and $224\times224$ resolution and apply color jitter during training.
We use the optimizer, learning rate, weight decay, and trainable modules listed in Table~\ref{tab:n16_training_parameters}.

All methods use the same source demonstrations, initial layouts, and controller limits.
At deployment, all policies use four denoising steps per inference call and execute 16-action chunks at a control frequency of 10~Hz.
All acceleration methods split intervals at gripper-command transitions. Only \ours{} applies the finger-motion cap.
DemoSpeedup estimates action entropy using the Original policy for the same task.
VOLT labels the external-camera video of each demonstration at 10 fps using the published instruction and sampling settings.
VOLT and DemoSpeedup use 1$\times$ for maintain-speed segments and 2$\times$ for speed-up segments on Franka.
Depending on the task, success requires a three-cup stack, the middle cup in the basket, or the chicken on the plate.
We exclude reset time from the mean completion time for successful trials.

\section{Additional Analysis}
\label{app:additional_results}

\subsection{Data Reference and Context Groups}
\label{app:analysis_reference}
For Fig.~\ref{fig:decision_data}\subref{fig:decision_context}, Uniform ($1.92\times$) produces 161,831 steps under the same retiming settings as \ours{} (159,221 steps), a difference of 1.64\%.
With the finger-motion cap fixed at 1.5, the compression ratios are 1.690 for Uniform and 1.718 for \ours{}, and their overall LIBERO ES values are 1.212 and 1.257, respectively.

For each group in Fig.~\ref{fig:decision_data}\subref{fig:decision_context}, we compute the relative change in output count using the Uniform count for that group as the denominator.
Transition windows contain a gripper-command change.
Adjacent windows border a transition window but contain no change, and Other covers the remaining windows.
The clearance and command groups overlap.
The nine windows with unknown clearance contribute only to the command groups.

\subsection{Candidate Comparison}
\label{app:decision_data_analysis}
\begin{table}[t]
  \centering
  \caption{Changes in training data for three candidate factors in window w7 of demonstration 650.
  Command peak denotes the highest requested linear speed. Path length measures the distance traveled between successive selected observations.}
  \label{tab:decision_w7_comparison}
{\small
  \setlength{\tabcolsep}{3pt}
  \renewcommand{\arraystretch}{1.05}
  \begin{tabular*}{\linewidth}{@{\extracolsep{\fill}}lccc@{}}
    \toprule
    \multicolumn{4}{@{}l@{}}{w7: open clearance; recorded path 143.4 mm} \\
    \midrule
    Candidate factor & 1.0 & 1.5 & 2.0 \\
    \midrule
    Output steps & 16 & 11 & 8 \\
    Command peak (m/s) & 0.986 & 1.592 & 1.821 \\
    Max. path length (mm) & 13.2 & 22.5 & 25.2 \\
    \bottomrule
  \end{tabular*}
}
\end{table}

Table~\ref{tab:decision_w7_comparison} compares the candidate retiming effects for window w7 in Fig.~\ref{fig:decision_data}\subref{fig:decision_episode}.
A factor of 2.0 removes three more steps than 1.5, but increases both the command peak and the maximum distance traveled between successive selected observations.
The term filter withholds the arrangement description for this window, but the remaining visual fields identify open clearance and list the bowl and container as obstacles.

We select the example from Goal task 6, where \ours{} has the largest success advantage over the Uniform policy in Appendix~\ref{app:analysis_reference}.
Eligible demonstrations contain 8 to 20 windows, both opening and closing gripper transitions, and distinct candidate and ablation outputs.
Of the two eligible demonstrations, we select demonstration 650, which has more differences in selected source indices from Uniform.

\end{document}